\documentclass{article}
\usepackage{ijcai25}
\usepackage{CJK}
\usepackage{times}
\usepackage{soul}
\usepackage{url}
\usepackage[hidelinks]{hyperref}
\usepackage[utf8]{inputenc}
\usepackage[small]{caption}
\usepackage{graphicx}
\usepackage{amsmath, amssymb}
\usepackage{amsthm}
\usepackage{booktabs}
\usepackage{algorithm}
\usepackage{algorithmic}
\usepackage[switch]{lineno}

\hypersetup{
    colorlinks=true,
    linkcolor=black,
    filecolor=black,      
    urlcolor=black,
    citecolor=black,
}

\title{Explainable Graph Neural Networks via Structural Externalities}

\author{
Lijun Wu$^1$
\and
Dong Hao$^{1,2}$\thanks{Corresponding author
}\And
Zhiyi Fan$^1$\\
\affiliations
$^1$SCSE, University of Electronic Science and Technology of China\\
$^2$AI-HSS, University of Electronic Science and Technology of China\\
\emails
lijunwu055@gmail.com, haodong@uestc.edu.cn,\\ fanzhiyi.027@gmail.com}

\begin{document}
\maketitle
\footnotetext{The full version of this paper and source code are at \href{https://github.com/wlj55/GraphEXT}{\textcolor{blue}{this link}}.}

\begin{abstract}

Graph Neural Networks (GNNs) have achieved outstanding performance across a wide range of graph-related tasks. However, their "black-box" nature poses significant challenges to their explainability, and existing  methods often fail to effectively capture the intricate interaction patterns among nodes within the network. In this work, we propose a novel explainability framework,  GraphEXT, which leverages cooperative game theory and the concept of social externalities. GraphEXT partitions graph nodes into coalitions, decomposing the original graph into independent subgraphs. By integrating graph structure as an externality and incorporating the Shapley value under externalities, GraphEXT quantifies node importance through their marginal contributions to GNN predictions as the nodes transition between coalitions. Unlike traditional Shapley value-based methods that primarily focus on node attributes, our GraphEXT places greater emphasis on the interactions among nodes and the impact of structural changes on GNN predictions. Experimental studies on both synthetic and real-world datasets show that GraphEXT outperforms existing baseline methods in terms of fidelity across diverse GNN architectures , significantly enhancing the explainability of GNN models.

\end{abstract}

\section{Introduction}

With the rapid advancement of deep learning, Graph Neural Networks (GNNs) have extended deep learning techniques to graph data. By efficiently leveraging structural information in graphs, GNNs have achieved remarkable performance in downstream tasks such as node classification~\cite{kipf2016semi,velivckovic2017graph} and graph classification~\cite{xu2018powerful,gao2019graph}. However, the reliance on complex and nonlinear functions renders GNN predictions challenging to explain intuitively. This limitation undermines the fairness and trustworthiness of GNNs, restricting their broader adoption in critical applications.

Compared to deep learning models in text and image domains, the explainability of GNNs remains underexplored. Existing studies, such as GNNExplainer~\cite{ying2019gnnexplainer}, PGExplainer~\cite{luo2020parameterized}, and SubgraphX~\cite{yuan2021explainability}, attempt to explain GNN predictions by identifying subsets of nodes, edges, or features that contribute most significantly to the results. However, these approaches exhibit notable limitations. Some methods, such as GraphLime~\cite{huang2022graphlime}, focus solely on node features while ignoring graph structural information. Others, despite considering structural elements like edges or subgraphs, often lack a theoretically grounded analysis of how graph structures influence GNN predictions. We have noticed that graph structure, as the core attribute of graph data, is directly reflected in the working principles of GNNs. We argue that the structure not only determines the interaction patterns among nodes but also has a profound impact on the GNN models' predictions, making structure-based explanations more intuitive and theoretically grounded.

In this work, we propose a novel GNN explanation method  GraphEXT, which models the intrinsic nature of graph structure. We introduce the concept of externality from economics into graph data modeling. Externality is an important concept in economics that refers to the impact, either positive or negative, of an economic agent's activity (or feature) on other agents not involved in that activity (or not having that feature).  Externalities arise whenever the actions of one economic agent directly affect another economic agent. Depending on the nature of the impact, externalities can be either positive or negative.
The positive externality occurs when an economic agent's activities benefit others, e.g. a person investing in education improves his or her skills and at the same time generates higher productivity for society.  In such cases, the benefits to society (society can be treated as a `mechanism' in economics) outweigh the benefits to the individual, and the market may undervalue the individual's education activity. On the contrary, negative externality occurs when the activities of one economic agent adversely affect others, e.g. the emission of pollutants from a factory causes deterioration of air quality of the society, etc.

Inspired by the economic externality, we can better capture the structural externalities' influence on GNN predictions and the interaction patterns among nodes. Our intuition is that, we can treat the GNN mechanism as the `society in economics', while the nodes and edges which are not enclosed by GNN can cause externality. In addition, the edges among the enclosed nodes also cause externalities to the GNN. The impact of non-enclosed nodes/edges could be either positive or negative.
With this inspiration of externality, to quantify the importance of nodes in these interactions, we employ the Shapley value~\cite{kuhn1953contributions} under externalities, calculating each node's marginal contribution as a measure of its impact. Additionally, we develop an efficient sampling method to compute the Shapley value, using it as an importance metric to identify key nodes for explanation. Compared to existing Shapley value-based GNN explanation methods, GraphEXT innovatively integrates economic theory for rigorous graph structure modeling, providing a more natural representation of its importance. This enables our approach to more effectively capture critical information within the graph structure. Extensive results demonstrate that our GraphEXT can identify key features in input graphs and consistently outperform baseline methods in explanation fidelity. These explanations not only enhance our understanding of GNNs but also inform improvements in model architectures, further boosting their performance.

\subsection{Related Works}

With the widespread adoption of GNNs, their explainability has garnered increasing attention, leading to the development of various explanation methods. Based on different perspectives, these methods can be broadly categorized into instance-level and model-level approaches \cite{yuan2022explainability}. Instance-level methods focus on explaining the behavior of a predictive model for each input graph by identifying relevant deep features.

For instance-level methods, gradient/feature based methods (e.g., SA~\cite{zeiler2014visualizing}, Guide BP~\cite{springenberg2014striving}, and GradCAM~\cite{pope2019explainability}) extend traditional explanation techniques for deep learning models to graph data by leveraging gradient changes in feature maps to assess input feature importance. While these methods are simple and efficient, they are prone to gradient saturation, which can hinder their ability to identify critical features. Decomposition based methods (e.g., LRP~\cite{baldassarre2019explainability}, Excitation BP~\cite{pope2019explainability}) allocate predictive scores layer by layer back to the input features, quantifying their importance. However, these methods often overlook the effects of activation functions, potentially misestimating nonlinear contributions. Proxy based methods (e.g., GraphLime~\cite{huang2022graphlime} and RelEX~\cite{zhang2021relex}) fit explainable models to sampled subsets of input graph data to infer feature importance. A notable limitation of these methods lies in their relatively low reliance on structural information, which may reduce their ability to capture node and edge interactions within the graph.

perturbation based methods are the most popular at the instance level. These methods infer the importance of nodes and edges by introducing slight perturbations to the input graph and observing the changes in the model’s predictions. GNNExplainer~\cite{ying2019gnnexplainer} maximizes mutual information to generate masks for edges and features, providing explanations for the predictions. PGExplainer~\cite{luo2020parameterized} trains a parameterized mask predictor to learn the importance distribution of edges and generates explanations based on their impact on the predictions. SubgraphX~\cite{yuan2021explainability} employs Monte Carlo tree search and uses the Shapley value as a scoring function to quantify the contribution of subgraphs, offering subgraph-level explanations. ZORRO~\cite{funke2022zorro} adopts a greedy algorithm to iteratively select nodes or node features by evaluating the consistency between new predictions and the original model predictions, but it may lead to suboptimal local explanations.

In contrast, model-level methods aim to directly explain the behavior of the entire GNN model without focusing on specific input instances. XGNN~\cite{yuan2020xgnn} trains a graph generator to produce graphs that contribute maximally to a target prediction label, where the generated graphs are considered as global explanations. ProtGNN~\cite{zhang2022protgnn} combines prototype learning with GNNs by embedding a conditional subgraph sampling module, selecting the parts of the input graph most similar to prototype subgraphs as explanations. ie-HGCN~\cite{yang2021interpretable}, designed for heterogeneous networks, introduces a hierarchical aggregation architecture at the object and type levels, extracting useful meta-paths for each object from all possible meta-paths to generate layered explanations.

\section{Preliminaries}

\subsection{Graph Neural Networks}

Formally, a graph $G$ with $n$ nodes and $d$-dimensional node features can be defined as $(X, A)$, where $X \in \mathbb{R}^{n\times d}$ represents the node feature matrix, and $A \in \mathbb{R}^{n\times n}$ denotes the adjacency matrix. If there exists an edge from node $i$ to node $j$, then $A_{ij}=1$; otherwise, $A_{ij}=0$. For node $v_j$, its feature is represented by the $j$-th row of $X$, denoted as $x_j$.

With the development of deep learning, numerous graph neural network architectures, such as GCN~\cite{kipf2016semi}, GAT~\cite{velivckovic2017graph}, and GIN~\cite{xu2018powerful}, have been proposed to address different graph tasks. The core of GNNs lies in the message-passing mechanism, where each node's new representation is learned by aggregating messages propagated from its neighbors. A general message-passing layer can be expressed as: \begin{align} 
x_i^{(k)}=\gamma^{(k)}\big(x_i^{(k-1)}, \Box_j\phi^{(k)}(x_i^{(k-1)}, x_j^{(k-1)})\big),\nonumber
\end{align} 
where $k$ denotes the layer index. The message-passing process consists of three steps: (i) The propagation step. An MLP $\phi$ computes the message passed along the edge from $v_j$ to $v_i$. (ii) The aggregation step. The aggregation function $\Box$ combines messages from all neighbors of $v_i$. (iii) The update step. Another MLP $\gamma$ updates the node representation $x_i^{(k)}$ by incorporating the aggregated message and the previous node representation. The final layer $L$ of a GNN outputs the node representation $x_i^{(L)}$, which serves as the final node embedding and is utilized in various downstream tasks.

\subsection{Shapley value with externalities}

A cooperative game can be represented as a tuple $(N, \mathcal{V})$, where $N$ denotes the set of players, and $\mathcal{V}$ represents the value function. Assuming there are $n$ players in the game, $N = {1, 2, \ldots, n}$ represents the set comprising all players. Any subset $S \subseteq N$ is referred to as a coalition. If a set of coalitions ${S_1, \ldots, S_k}$ satisfies $\bigcup_{i=1}^k S_i = N$ and $\forall i \neq j, S_i \cap S_j = \emptyset$, it is referred to as a coalition structure.

In cooperative games with externalities, the payoff of a coalition $S$ is determined not only by its internal members but also by the coalition structure $P$ to which it belongs. The basic structure $(S, P)$ represents a specific coalition along with its associated coalition structure, and the set of all such structures constitutes $\mathcal{C}$. The value function $\mathcal{V}: \mathcal{C} \to \mathbb{R}$ defines the payoff of  coalition $S$ under the basic structure $(S, P)$.

The Shapley value~\cite{kuhn1953contributions} is a method in cooperative game theory that fairly allocates the total payoff of the game among players based on their contributions during the formation of coalitions. In cooperative games with externalities, the Shapley value focuses on the process where some players leave their original coalition in the structure $P$ and form a new coalition $S$, denoted as $P_{[S]} = {T \setminus S \mid T \in P} \cup {S}$. Given the value function $\mathcal{V}$ and any coalition structure $P$, the marginal contribution of player $i$ to coalition $S$ is defined as:
\begin{align}
    mc(S,i)=\mathcal{V}\left(S\cup\{i\},P_{[S\cup\{i\}]}\right)-\mathcal{V}\left(S,P_{[S]}\right).\nonumber
\end{align}

In the presence of externalities, the computation of the Shapley value is not unique. To unify the description of such algorithms, Skibski\cite{skibski2013shapley} artificially defines a weight function for player $i$, $\alpha_i: (S, P) \mid i \not\in S \to [0, 1]$. By iterating over all ordered coalitions and averaging their marginal contributions, the Shapley value for each player is obtained:
{\small \begin{align}
\varphi_i(\mathcal{V})=\frac{1}{n!}\sum_{\pi\in \Omega(n)}\sum_{P}\prod_{j\in N}\alpha_j(C_j^{\pi}, P_{[C_j^{\pi}]})\cdot mc(C_{i}^{\pi}, i)\label{SV}.
\end{align}}Here, $\Omega(n)$ is all permutations of length $n$, and $C_i^\pi$ denotes the set of players preceding player $i$ in permutation $\pi$.

\section{Proposed Method}

\begin{figure*}[t]
\centering
\includegraphics[width=0.9\linewidth, trim=0cm 1cm 0cm 3cm]{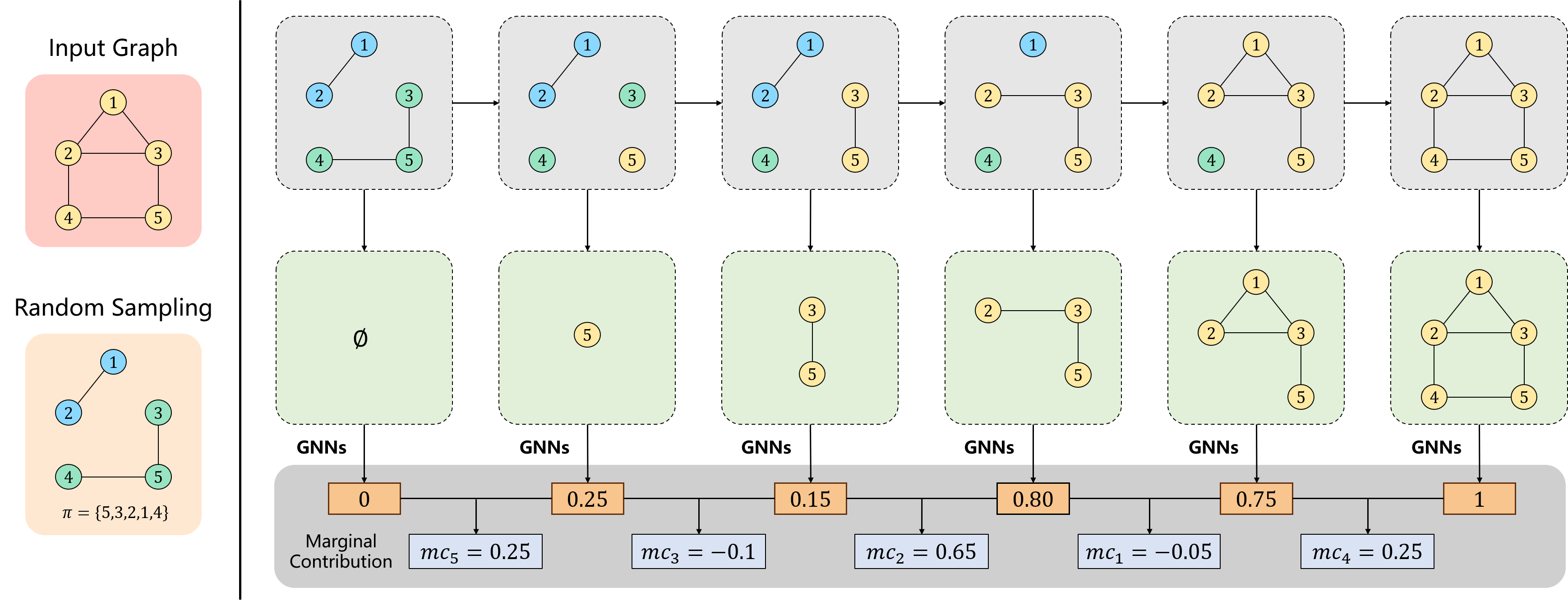}
\caption{An illustration of our proposed GraphEXT. \textbf{Left}: The input graph to be explained and the sampled tuple $(\pi, P)$ in one iteration. \textbf{Right}: During each sampling iteration, the original graph is partitioned into several connected components based on the coalition structure $P$. Nodes are sequentially transferred from their original coalitions to the coalition $S$ (marked as blue nodes) following the order defined by the permutation $\pi$. The subgraph corresponding to $S$ is then fed into the GNN to obtain the model's output. The marginal contribution of a node during this sampling process is calculated as the difference in GNN outputs before and after the node is added to $S$.}
\label{fig:framework}
\end{figure*}

\subsection{Explaining GNNs with Node Importance}

Without loss of generality, let $f(X, A)$ or $f(G)$ represent a graph neural network designed for graph classification tasks, where $G$ is the input graph, and $X, A$ denote its node feature matrix and adjacency matrix, respectively. As a black-box model, the architecture, parameters, and training data of $f$ are inaccessible; only the model's output corresponding to any given inputs is available. We illustrate our method using this network and subsequently discuss its generalization to node classification and link prediction tasks.

Assume the network comprises $L$ GNN layers followed by a graph classification function $g$. The network processes $X$ and $A$ as inputs and leverages the node representations $X^{(L)}$ learned by the $L$ layers as inputs to $g$. The classification function $g$ typically includes a readout operation, such as global average pooling, combined with a multi-layer perceptron serving as the classifier. For a $C$-class classification problem, the model outputs a conditional probability distribution $P(Y|G)$, where the random variable $Y$ spans $C$ labels ${1, \ldots, C}$, each representing the probability of the input graph belonging to a specific class. The final classification result corresponds to the label with the highest predicted probability. We use $f(G)_{y_i}$ to denote the model's predicted probability for class $y_i$, i.e., $P(Y = y_i | G)$.

Formally, given the input $X$ and $A$, the model predicts the label $y$. Let $\mathcal{G} = {G_1, \ldots, G_{|\mathcal{G}|}}$ be a set of candidate graphs intended to serve as explanations for the prediction. The task of explaining the prediction can be formulated as the following optimization problem:
\begin{align}
G^*=\mathop{\arg\max}_{G\in \mathcal{G}} f(G)_y.\nonumber
\end{align}

A method to generate the candidate set $\mathcal{G}$ is to enumerate all subsets $S \subseteq N$ and create a node mask $\mathbf{M_S}$ based on the following rule: if $v_i \in S$, the $i$-th entry of $\mathbf{M_S}$ is set to $1$; otherwise, it is set to $0$. Using the mask, a new graph $G' = (X \odot \mathbf{M_S}, A)$ is constructed, where $\odot$ denotes element-wise multiplication. Compared to the input graph, the adjacency matrix of $G'$ remains unchanged, while the node feature matrix is modified: for $v_i \in S$, the node features $x_i$ are retained; otherwise, $x_i = \mathbf{0}$. The new graphs corresponding to all subsets collectively form the candidate set $\mathcal{G}$.

However, the size of $\mathcal{G}$ and the complexity of its generation grow exponentially with the graph size, making it inefficient for model explanation. Node-level explanation methods typically quantify the importance of each node with a real value and select the most important nodes as the explanation result based on task requirements.

Most existing node-level explanation methods overlook the role of nodes in the graph structure and lack a solid theoretical foundation to evaluate the structural importance of nodes. We argue that the graph structure is a fundamental property of nodes, and explanations derived from structural features are more natural and theoretically grounded. To address this, we propose a method named GraphEXT, which models graph structure as an externality influencing GNN predictions and quantifies node importance through their interactions via the graph structure. Specifically, we introduce a graph-based game-theoretic framework incorporating Shapley value under externalities and adopt an efficient sampling strategy. The computed Shapley value serves as a quantitative measure of node importance, with the most significant nodes selected as the explanation result. Our approach is illustrated in Figure~\ref{fig:framework}.

\subsection{Externality of Graph Structure}
\label{section42}

As mentioned above, the quantification of node importance is based on the calculation of the Shapley value under externality defined in eq.~(\ref{SV}). The relevant definitions of cooperative game theory will be extended to the graph domain in this section. Given the graph $G=(X,A)$, let its node set be $V=\{v_1, \dots, v_n\}$, and define the set of players in the game as $N=V$. A subset of the node set, $S=\{v_{S_1}, \dots, v_{S_k}\}$, forms a coalition. When only considering the nodes in the coalition $S$, the graph $G$ is divided into several connected components, and the set of these connected components is defined as $S \mid G$: 
\begin{align} 
S\mid G = \{\{i \mid i \text{ and } j \text{ are connected in } S \text{ by } G\} \mid j \in S\}.\nonumber 
\end{align} 
Since the aggregation mechanism of GNN operates on each node and its neighboring nodes, the output of GNN for each connected component of the input graph is independent. Therefore, we have 
\begin{align} 
f(G_S) = \sum_{T \in S} f(G_T).\nonumber 
\end{align} 
Here, $G_S$ represents the new graph $(X, A \odot \mathbf{M_S}\mathbf{M_S}^T)$ formed by removing all nodes in $N \setminus S$ and the edges incident to them from $G$.

Based on Myerson's definition~\cite{myerson1977graphs,navarro2007fair}, we define $\mathcal{W}(T,G)$ as the value function of a maximal connected component $T$ in graph $G$, where its value is related to the internal structure of component $T$, meaning that the graph structure $G$ exists as an externality. In this problem context, it is naturally assumed that $\mathcal{W}(T, G) = f(G_T)$. When the internal structure of $G_T$ changes, even if the set $T$ remains the same, the value of $\mathcal{W}(T, G)$ must change accordingly. The above expression can be rewritten as: 
\begin{align} 
f(G_S) = \sum_{T \in S} \mathcal{W}(T, G_S).\nonumber 
\end{align}

Given the coalition structure $P$, the original graph is decomposed into a new graph $G_P$ based on the coalitions in $P$: 
\begin{align} G_P = \left(X, A \odot \big(\sum_{S \in P} \mathbf{M_S} \mathbf{M_S}^T\big)\right).\nonumber 
\end{align}

Based on the previous analysis, we can naturally define the value function of the current coalition game with the graph structure as an externality, $\mathcal{V}(S,P)$, as the output of GNN when the subgraph $G_S$ corresponding to coalition $S$ is input under the structure $G_P$: 
\begin{align} \mathcal{V}(S,P) = \sum_{T \in S} \mathcal{W}(T, G_P).\nonumber 
\end{align}

The algorithm flow for this step is shown in Algorithm~\ref{alg:alg1}. First, we identify all edges whose endpoints belong to a same coalition in $P$ and construct a subgraph of the original graph using these edges. The connected components of coalition $S$ in the graph are found using a breadth-first search. Summing their corresponding outputs in the GNN gives the value of $\mathcal{V}(S,P)$. The computed value function $\mathcal{V}(S,P)$, together with the player set $N$, jointly defines the coalition game model we use.

To quantify the importance of each node, we adopt the Shapley value calculation method based on externality as proposed in Macho-Stadler's work\cite{macho2007sharing}: 
\begin{align} 
\varphi_i(\mathcal{V}) = \sum_{(S,P) \in \mathcal{C}} \frac{\prod_{T \in P \setminus S} (|T|-1)!}{(|N|-|S|)!} \beta_i(S) \mathcal{V}(S,P), 
\label{ext}
\end{align}
where, 
$$\beta_i(S) = \left\{ \begin{aligned} &\frac{(|S|-1)!(|N|-|S|)!}{|N|!} & , i \in S \\ &-\frac{|S|!(|N|-|S|-1)!}{|N|!} & , i \in N \setminus S \end{aligned} \right..$$

The advantage of this calculation method is that it does not require explicit modeling of externality (such as assuming a linear relationship or some specific form of externality), making it suitable for graph neural networks with black-box properties. At the same time, it retains the efficiency, symmetry, and null-player properties of traditional Shapley values, ensuring fairness in the allocation algorithm theoretically. The computed Shapley values are directly used as the importance of each node. Since we use the GNN's prediction results as the value function for the coalition game when the graph structure changes, the physical meaning of the Shapley value leans more towards the contribution of nodes in the graph structure, offering a more fundamental explanation for the GNN's predictions.

\begin{algorithm}[tb]
\caption{Compute the Network Based Value Function}
\label{alg:alg1}
\textbf{Input}: Model $f(\cdot)$, number of nodes $n$, graph $G=(X,A)$, coalition $S$, coalition structure $P$\\
\textbf{Output}: Value of $\mathcal{V}(S, P)$
\begin{algorithmic}[1]
\FOR{$i = 1$ \textbf{to} $n$} 
\STATE $out_i \gets \{j\mid A_{ij}=1$ and $i,j$ belong to the same coalition $\}$
\ENDFOR
\STATE $components \gets []$
\FOR{$i = 1$ \textbf{to} $n$}
\IF{$i \in S$ \textbf{and} $i$ has not been assigned to a connected component}
\STATE $queue \gets [i], visited \gets \{i\}$
\WHILE{$queue$ is not empty}
\STATE Extract the first element $u$
\STATE $queue \gets queue +  (out_u\setminus visited)$
\STATE $visited \gets visited + out_u$
\ENDWHILE
\STATE $components \gets components + [visited]$
\ENDIF
\ENDFOR
\STATE \textbf{return} $\sum_{R\in components} f(G_R)$
\end{algorithmic}
\end{algorithm}

\subsection{Unbiased Sampling Method}

In practice, the exact calculation of the Shapley value requires considering all possible coalition structures. As the number of players increases, the computational complexity grows exponentially, so sampling methods are used to approximate the Shapley value.

As shown in eq.~(\ref{SV}), when calculating the Shapley value for player $i$, there is a weight coefficient before the marginal contribution $mc(S, i)$ for each coalition. It can be proven that:
\begin{align} 
\frac{1}{n!}\sum_{\pi \in \Omega(n)}\sum_{P}\bigg[\prod_{j \in N} \alpha_j(C_j^{\pi}, P_{[C_j^{\pi}]})\bigg] = 1.\nonumber 
\end{align}

Based on this result, when calculating the Shapley value for player $i$, we treat the weight coefficient as the probability of sampling the pair $(\pi, P)$. In the $m$-th sampling step, we sample a pair $(\pi, P)$ according to this probability and compute the marginal contribution $mc_m(i)$ for player $i$. Let the total number of sampling steps be $T$, and the average of the marginal contributions obtained from $T$ samples is taken as the estimated Shapley value $\hat{\varphi_i}(\mathcal{V})$ for player $i$. Since the sampling steps are independent, it can be proven that this sampling method is unbiased: 
\begin{align} 
\mathbb{E}[\hat{\varphi_i}(\mathcal{V})] & = \frac{1}{T} \sum_{m=1}^T \mathbb{E}[mc_m(i)] = \mathbb{E}[mc(i)] \nonumber \\ & = \frac{1}{n!} \sum_{\pi \in \Omega(n)} \sum_{P} \prod_{j \in N} \alpha_j(C_j^{\pi}, P_{[C_j^{\pi}]}) \cdot mc(C_{i}^{\pi}, i) \nonumber \\ & = \varphi_i(\mathcal{V}).\nonumber \end{align}

Based on Skibski's work \cite{skibski2013shapley}, the weight function corresponding to the algorithm in eq.~(\ref{ext}) is: \begin{align} \alpha_i(S,P) = \frac{\max(|P(i)| - 1, 1)}{n - |S|},\nonumber \end{align} where $P(i)$ represents the coalition in which player $i$ is located in the coalition structure $P$. Referring to Skibski's work, to satisfy this weight function, we use the Kunth Shuffle to generate two random permutations, one of which is used as $\pi$, and the permutation cycles from the other are extracted. Each cycle is treated as a coalition to generate $P$.

In the algorithm implementation, we do not independently sample $T$ permutations $\pi$ for each player and construct the corresponding coalition $C_i^{\pi}$. Instead, at each sampling step, we sample a tuple $(\pi, P)$ and iteratively move players out of their original coalitions into coalition $S$ according to the order of the permutation $\pi$, with $S$ initially being an empty coalition. It can be observed that after adding player $\pi(i)$, the set $S$ is equivalent to the coalition $C_{\pi(i)}^{\pi}$. This process continues until all players are added to $S$, and the marginal contribution of each player is computed for this sampling step. By averaging the marginal contributions over $T$ samples, the Shapley value for each player is estimated. The specific algorithm for estimating Shapley values is shown in Algorithm~\ref{alg:alg2}.

\begin{algorithm}[tb]
\caption{Shapley Value Sampling and Estimation}
\label{alg:alg2}
\textbf{Input}: Value function $\mathcal{V}(S, P)$, number of samples $T$ \\
\textbf{Output}: Shapley value for each player $\varphi_z \mid z\in V$

\begin{algorithmic}[1]

\STATE \textbf{for all $z\in v$ do $\varphi_z\gets 0$ end for}

\FOR{$i = 1$ \textbf{to} $T$}
    \STATE $\pi,A \gets$ two random permutations generated using Knuth Shuffle
    \STATE $S \gets \emptyset$
    \STATE Construct a directed graph with $n$ nodes and $n$ directed edges based on permutation $A$, where the $i$-th edge points from node $i$ to node $A(i)$
    \STATE Identify all permutation cycles in the graph, with each cycle representing a coalition, and construct the initial partition $P$
    
    \FOR{$j = 1$ \textbf{to} $n$}
        \STATE $\mathcal{V}_{\text{before}} \gets \mathcal{V}(S, P)$
        \STATE $S\gets S\cup\{\pi(j)\},P\to P_{[S]}$
        \STATE $\mathcal{V}_{\text{after}} \gets \mathcal{V}(S, P)$
        \STATE  $\varphi_{\pi(j)} \gets \varphi_{\pi(j)}+ \frac{1}{T}(\mathcal{V}_{\text{after}} - \mathcal{V}_{\text{before}})$
    \ENDFOR
\ENDFOR

\STATE \textbf{return} $\varphi_z \mid z\in V$
\end{algorithmic}
\end{algorithm}

\begin{figure*}[t]
\centering
\includegraphics[width=0.95\linewidth, trim= 0cm 0.5cm 0cm 1.5cm]{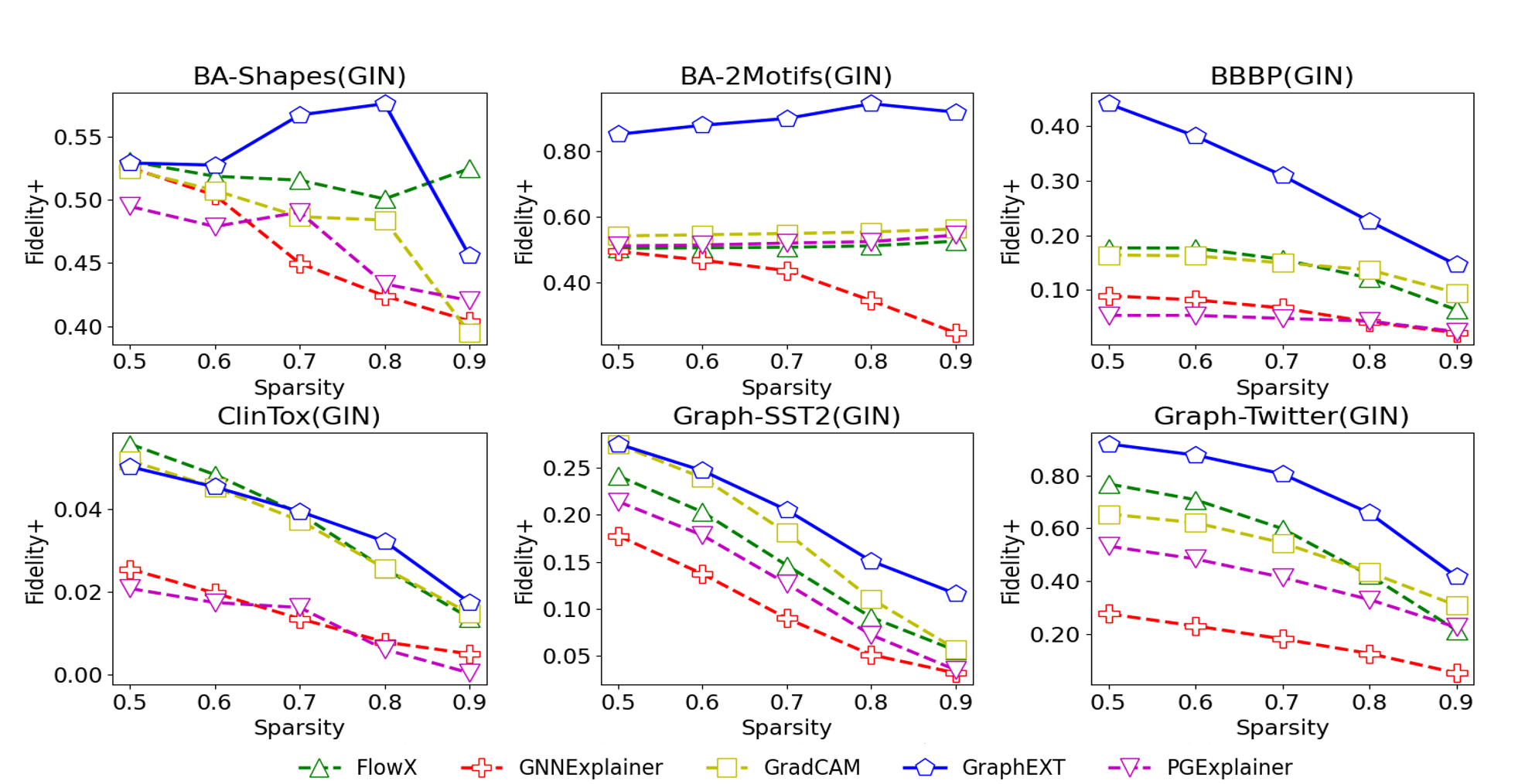}
\caption{Fidelity+ values on 6 datasets with GINs under different Sparsity levels, higher Fidelity+ indicates better performance.}
\label{fig:res2}
\end{figure*}

\subsection{Model Extensions}

The above sections describes our proposed method using the graph classification task as an example. It is worth noting that this method can be adapted to explain node classification and link prediction models with appropriate adjustments. For node classification models, given a graph $G$, we aim to explain the model $f(\cdot)$'s prediction for the target node $v_i$. Since an $L$-layer graph neural network only considers the $L$-hop neighborhood of $v_i$, we extract the $L$-hop subgraph centered on $v_i$ to reduce computational complexity. As the target of the model's prediction, $v_i$ must be included in every extracted connected component; otherwise, the model cannot classify $v_i$. Based on this, we handle $v_i$ separately and modify the set of players in the coalition game model proposed in Section~\ref{section42} to $N\setminus\{v_i\}$. For a maximal connected component $T$ formed by coalition $S$, if $T\cup{v_i}$ remains connected (i.e., there exists a node in $T$ connected to $v_i$), the value function is defined as $\mathcal{W}(T,G)=f(G_{T\cup\{v_i\}})$. Otherwise, $\mathcal{W}(T,G)=0$. When extracting the top-$k$ important nodes as the explanation, we select the target node $v_i$ and the top $k-1$ nodes ranked by Shapley value. For link prediction tasks, the goal is to explain the model's prediction for the existence probability of an edge $(v_i, v_j)$. We extract the $L$-hop subgraph around $v_i$ and $v_j$ and treat the target node pair ${v_i, v_j}$ separately. Correspondingly, we adjust the definitions of the characteristic function and the value function to adapt to this task.

\begin{figure*}[t]
\centering
\includegraphics[width=0.95\linewidth, trim= 0cm 0.5cm 0cm 1.5cm]{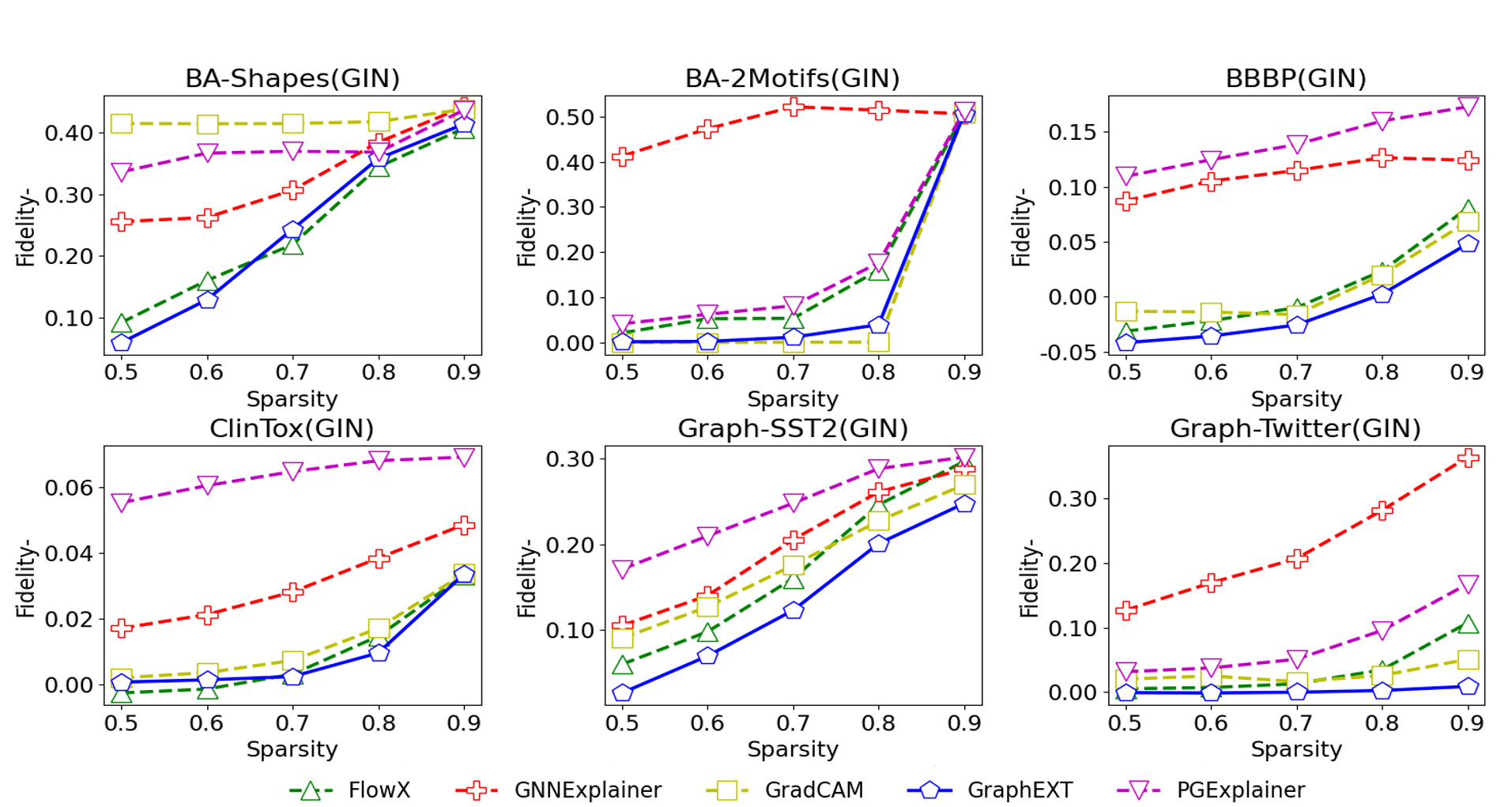}
\caption{Fidelity- values on 6 datasets with GINs under different Sparsity levels, lower Fidelity- indicates better performance.}
\label{fig:res4}
\end{figure*}

\section{Experimental Studies}

\subsection{Datasets and Baselines}

We evaluated the effectiveness of GraphEXT using six datasets from node classification and graph-level classification tasks. The statistical details of these datasets are shown in Table~\ref{tab:tab1}. These datasets include synthetic datasets, sentiment graph datasets, and biological datasets.  

 BA-Shapes~\cite{ying2019gnnexplainer} is designed for node classification tasks, it is based on a Barab\'asi-Albert graph with added house-like patterns. Nodes are categorized into four classes based on whether they are part of the house pattern and their positions within the pattern.

 BA-2Motifs~\cite{luo2020parameterized} is used for graph classification tasks, it is also based on a Barab\'asi-Albert graph with two types of house-like patterns added. Graphs are classified into two categories based on the patterns.

 Graph-SST2 and Graph-Twitter~\cite{yuan2022explainability} are used for graph classification tasks, these datasets transform textual data into graph representations, where nodes are words and edges are word associations. Each graph is classified based on its sentiment polarity (positive or negative).

 BBBP and ClinTox~\cite{wu2018moleculenet} are designed for graph classification tasks, each graph in these datasets represents a molecule. Nodes are atoms, edges represent chemical bonds, and graph labels are determined by the molecule's chemical functions or properties.

\begin{table}[tb]
    \centering
    \begin{tabular}{cccc}
        \hline
        Datasets  & Graphs & Nodes(avg) & Classes \\
        \hline
        BA-Shapes     & 1     & 700 & 4 \\
        BA-2Motifs  & 1000     & 25 & 2\\
        BBBP & 2039     & 24.06 & 2 \\
        ClinTox & 1478     & 26.16 & 2 \\
        Graph-SST2 & 70042 & 10.20 & 2\\
        Graph-Twitter & 6940 & 21.10 & 2\\
        \hline
    \end{tabular}
    \caption{Statistics and properties of six datasets}
    \label{tab:tab1}
\end{table}

In our experiments, we trained on all datasets using three-layer GCN~\cite{kipf2016semi} and GIN~\cite{xu2018powerful} models. The model with the highest accuracy on the test set was selected as our final model, and all models were trained to achieve competitive accuracy. Subsequently, we chose several state-of-the-art methods for graph structure explanation as baselines, including FlowX~\cite{gui2023flowx}, GNNExplainer~\cite{ying2019gnnexplainer}, GradCAM~\cite{pope2019explainability}, and PGExplainer~\cite{luo2020parameterized}. The datasets and baseline implementations were based on the DIG Library~\cite{liu2021dig}.

\subsection{Experimental Results on Datasets}

We adopted Fidelity+$^{prob}$, Fidelity-$^{prob}$~\cite{yuan2021explainability,yuan2022explainability}, and Sparsity~\cite{yuan2022explainability} as evaluation metrics. Given a model $f(\cdot)$, a graph to be explained $G_i=(X_i,A_i)$, the model’s classification result $y_i$, and the node mask $\mathbf{M_i}$ generated by the explanation method, these metrics are defined as follows:
\begin{align}
\text{Fidelity+}^{prob}=\frac 1N \sum_{i=1}^{N} f(G)_{y_i}-f(G^{1-\mathbf{M_i}})_{y_i},\nonumber
\end{align}
\begin{align}
\text{Fidelity-}^{prob}=\frac 1N \sum_{i=1}^{N} f(G)_{y_i}-f(G^{\mathbf{M_i}})_{y_i},\nonumber
\end{align}
\begin{align}
    \text{Sparsity}=\frac 1N \sum_{i=1}^N (1-\frac{m_i}{M_i}).\nonumber
\end{align}

Here, $G^{\mathbf{M}}$ represents the new graph $(X \odot M, A)$ generated by retaining only the node features where $\mathbf{M}$ equals 1, $m_i=\|\mathbf{M}\|_1$ denotes the number of important nodes selected, and $M_i$ indicates the total number of nodes in the graph. We evaluated our proposed GraphEXT on both synthetic and real-world datasets. For each dataset, we conducted quantitative calculations of the Fidelity metrics using test samples on trained GCN and GIN models to demonstrate the effectiveness of our method. The experimental results and analyses for the GIN model are presented in the main text. The results and related discussions for the GCN model are provided in the supplementary material for further reference.

We first conducted a quantitative analysis of Fidelity+ and Sparsity for different explanation methods. An ideal explanation method should accurately capture the critical features upon which the model's predictions are based. When these features are removed, the model's prediction probability should change significantly. Fidelity+ measures the probability change in the model's classification of the original input after removing the identified important features. A higher Fidelity+ score indicates that the method has identified more essential features. Sparsity evaluates the proportion of significant features to the total features in the original input, where a higher Sparsity score means fewer features are selected as critical. Ideally, a method should achieve higher Fidelity+ while maintaining lower Sparsity. Therefore, we compared the Fidelity+ scores of various methods under controlled Sparsity levels. As shown in Figure~\ref{fig:res2}, our method outperforms baseline methods significantly at different Sparsity levels across most datasets. Even in datasets where it does not outperform, its performance is comparable to the best baseline methods.

Subsequently, we compared Fidelity- and Sparsity for different explanation methods. Fidelity- measures the probability change in the model's classification of the original input when only the important features are retained. Lower Fidelity- values indicate that the method can better preserve critical features. As shown in Figure~\ref{fig:res4}, our method consistently achieves significantly lower Fidelity- than baseline methods in most datasets. 

These experimental results demonstrate the robustness and generalization capabilities of our method across diverse datasets and model architectures. Furthermore, our method excels in explaining sentiment graph datasets, likely because such datasets often rely on a few sentiment-laden words to determine the overall sentiment. our method effectively quantifies the marginal contributions of these words within the graph structure, enabling more accurate explanations.

\section{Conclusion}

Graph Neural Networks (GNNs) are often regarded as black boxes, and improving their interpretability is a crucial research challenge. To address this issue, we propose GraphEXT, which introduces the concept of externality and integrates coalition game theory into the task of explaining GNNs. Our approach innovatively analyzes the impact of node transitions across different subgraphs on the model's predictions. Specifically, we leverage the Shapley value to quantitatively assess the importance of each node, delivering more interpretable results. Experimental findings demonstrate that GraphEXT consistently achieves superior interpretability across multiple synthetic and real-world datasets, as well as diverse GNN architectures. We believe this study not only offers a novel perspective for advancing GNN interpretability but also provides valuable insights for future research on graph-based modeling and explanation.
\section*{Acknowledgement}
The authors gratefully acknowledge the reviewers for their valuable comments and suggestions.
This research was supported by the National Natural Science Foundation of China (Grant No. 71601029) and the Sichuan Science and Technology Program (Grant No. 2025HJPJ0006).
Lijun Wu is also supported by the student research project provided by the YingCai Honors College of the University of Electronic Science and Technology of China.
\bibliographystyle{named}
\bibliography{ijcai25}

\clearpage
\appendix
\section{Complexity Analysis}
Without loss of generality, we take GCN as an example to analyze the complexity of our proposed GraphEXT. Let the number of nodes and edges in the input graph be denoted as $n$ and $m$, respectively, and let the dimensionality of the feature vector for each node be $d$. The complexity of the two proposed algorithms can be computed as follows:

(1) Value Function Computation: After generating the coalition structure $P$, we use breadth-first search  to divide the input graph into multiple independent subgraphs, with a time complexity of $\mathcal{O}(m)$. For each partitioned subgraph, we input it into the GCN model $f(\cdot)$ and take the output value as the value function of the corresponding subgraph. Let $m_i$ denote the number of edges in subgraph $i$. The time complexity of this step is $\mathcal{O}(\sum_i m_id) \le \mathcal{O}(md)$.

(2) Shapley Value Sampling and Estimation: During each sampling step, it takes $\mathcal{O}(n)$ complexity to generate the coalition structure of nodes and random permutations. Following the permutation order, we compute the value function of each node according to the method in (1), which requires a total time complexity of $\mathcal{O}(nmd)$. Given the number of samples $T$, the total time complexity of the algorithm is $\mathcal{O}(Tnmd)$. Generally, we set $T=100$, and the estimated value of the Shapley Value becomes stable.

\section{Efficiency Studies}

\begin{table}[b]
    \centering
    \begin{tabular}{ccc}
        \hline
        Method & Time/s & Fidelity+  \\
        \hline
        GraphEXT & 39.3 $\pm$ 17.1 & \textbf{0.78 }
        \\SubgraphX & 73.5 $\pm$ 39.2 & 0.51
        \\FlowX & 21.0 $\pm$ 4.1 & 0.68 \\GNNExplainer & 1.65 $\pm$ 0.64 & 0.28 \\GradCAM & 0.07 $\pm$ 0.002 & 0.64
        \\PGExplainer & \textbf{0.04 $\pm$  0.001}(Training 698s) & 0.37 
        \\
        \hline
    \end{tabular}
    \caption{Efficiency study of different methods}
    \label{tab:tab3}
\end{table}

Finally, we study the efficiency of the proposed GraphEXT. We selected 50 graphs from the Graph-Twitter dataset, with each graph containing an average of 21.9 nodes and 39.6 edges. The algorithms used in previous experiments were tested to evaluate their runtime performance. Additionally, to compare our method with other algorithms that also utilize Shapley value and sampling, we included a comparison with SubgraphX. The experimental results are presented in Table~\ref{tab:tab3}.

The experimental results demonstrate that, although GraphEXT introduces higher complexity due to the sampling algorithm, it achieves a significant improvement in fidelity. Furthermore, compared to the runtime of SubgraphX, the efficiency of our sampling algorithm is also evident. In summary, we believe that the trade-off between efficiency and fidelity achieved by GraphEXT is acceptable.

\section{Additional Experimental Results}

\begin{figure*}[tb]
\centering
\includegraphics[width=0.95\linewidth, trim= 1cm 0.5cm 1cm 1.5cm]{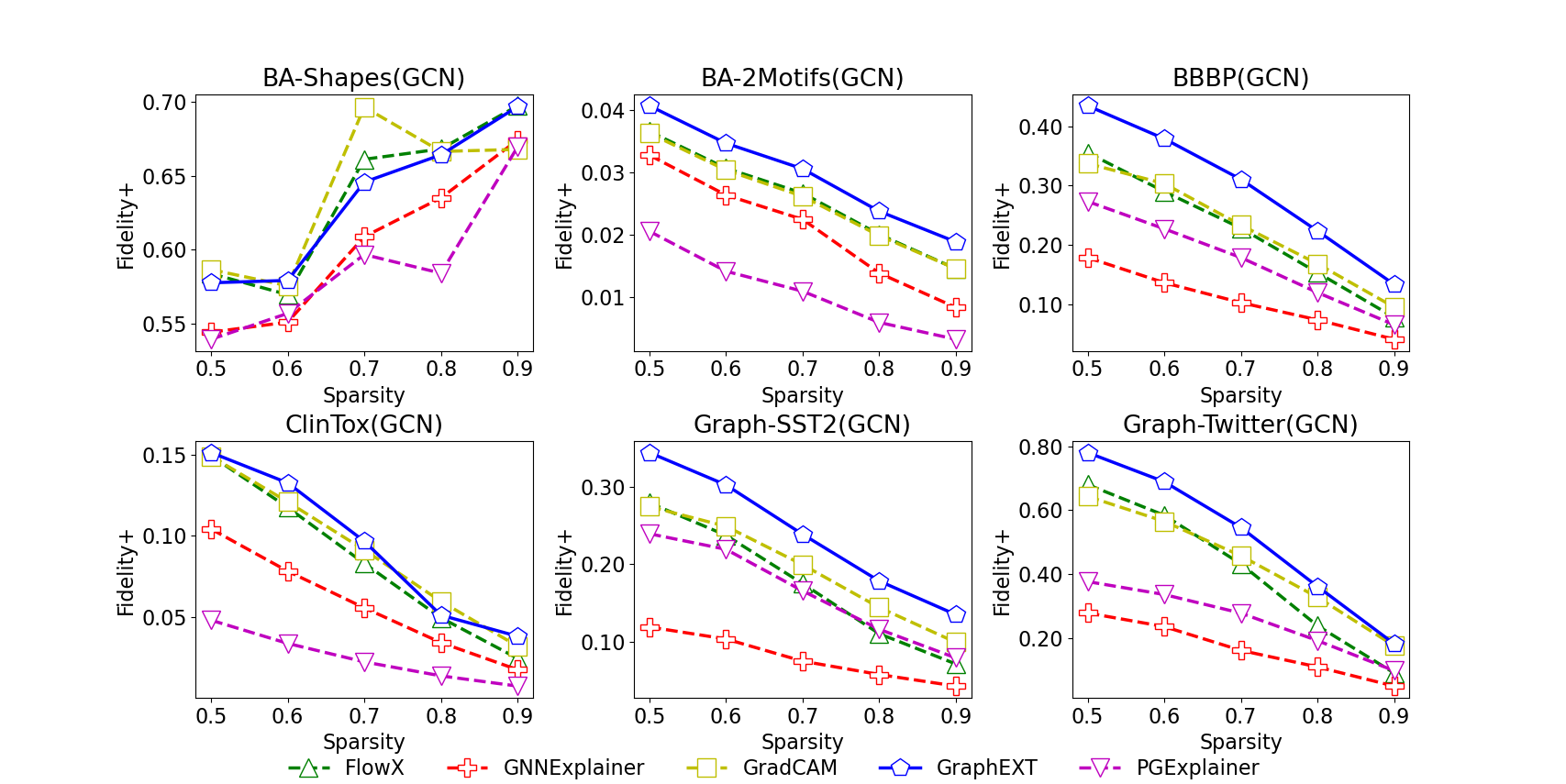}
\caption{Fidelity+ values on 6 datasets with GCNs under different Sparsity levels, higher Fidelity+ indicates better performance.}
\label{fig:res5}
\end{figure*}
\begin{figure*}[tb]
\centering
\includegraphics[width=0.95\linewidth, trim= 1cm 0.5cm 1cm 1.5cm]{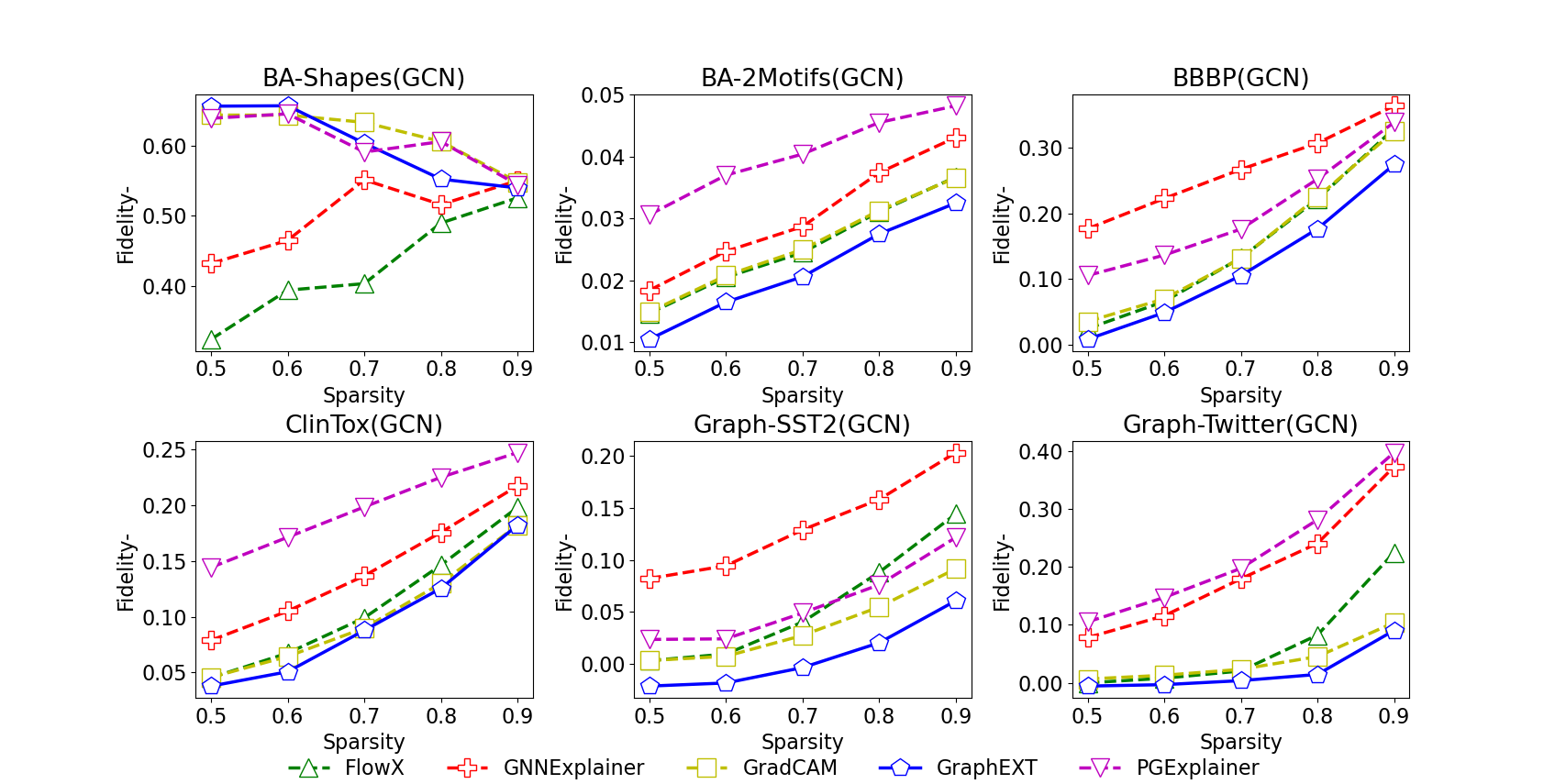}
\caption{Fidelity- values on 6 datasets with GCNs under different Sparsity levels, lower Fidelity- indicates better performance.}
\label{fig:res6}
\end{figure*}

In this section, we compare the performance of different algorithms in explaining the prediction results of GCN. The experimental results are shown in Figures~\ref{fig:res5} and~\ref{fig:res6}.

As observed, our proposed GraphEXT outperforms the other methods on the majority of datasets. However, its performance on the BA-Shapes dataset degrades under low sparsity conditions. This may be attributed to the presence of high-degree nodes in the target graph, which can lead to a large number of redundant subgraphs or isolated node pairs during coalition partitioning. These issues make it more challenging to accurately estimate the importance of each node.





\end{document}